%% file: iclr2024_conference.tex
\documentclass{article} 
\usepackage{iclr2024_conference,times}

\input{math_commands.tex}

\usepackage{hyperref}
\usepackage{hyperref}

\usepackage{siunitx}
\usepackage{booktabs} 
\usepackage[capitalise]{cleveref}
\usepackage{svg}
\usepackage{subcaption}
\usepackage{caption}
\usepackage{wrapfig}

\usepackage{tikz}
\usetikzlibrary{positioning, shapes, arrows, fit, backgrounds}

\DeclareSIUnit{\million}{\text{million}}
\DeclareSIUnit{\billion}{\text{billion}}

\title{Global Vegetation Modeling with \newline Pre-trained Weather Transformers}

\author{Pascal Janetzky, Florian Gallusser, Simon Hentschel, Andreas Hotho,  Anna Krause\\
Data Science Chair, 
Center for Artificial Intelligence and Data Science (CAIDAS), \\
University of Würzburg \\
\texttt{\{janetzky,gallusser,hentschel,hotho,}
\and
\texttt{anna.krause\}@informatik.uni-wuerzburg.de} \\
}

\iclrfinalcopy
\begin{document}

\maketitle

\begin{abstract}
Accurate vegetation models can produce further insights into the complex interaction between vegetation activity and ecosystem processes.
Previous research has established that long-term trends and short-term variability of temperature and precipitation affect vegetation activity.
Motivated by the recent success of Transformer-based Deep Learning models for medium-range weather forecasting, we adapt the publicly available pre-trained FourCastNet to model vegetation activity while accounting for the short-term dynamics of climate variability.
We investigate how the learned global representation of the atmosphere's state 
can be transferred to model the normalized difference vegetation index (NDVI).
Our model globally estimates vegetation activity at a resolution of \SI{0.25}{\degree} while relying only on meteorological data. 
We demonstrate that leveraging pre-trained weather models improves the NDVI estimates compared to learning an NDVI model from scratch. 
Additionally, we compare our results to other recent data-driven NDVI modeling approaches from machine learning and ecology literature. 
We further provide experimental evidence on how much data and training time is necessary to turn FourCastNet into an effective vegetation model.
Code and models will be made available upon publication.
\end{abstract}

\section{Introduction}
Environmental changes affect the dynamics of terrestrial vegetation, which is involved in controlling water, energy and CO\textsubscript{2} fluxes \citep{richardson2013climate}, and is thus crucial for providing ecosystem services such as food, fiber and fuel \citep{piao2020}.
Hence, a profound understanding of the complex interplay of climate system variables and vegetation changes is desirable to achieve sustainable ecological management.

Previous studies have shown that observed changes in vegetation can be attributed to both long-term and short-term changes in temperature and precipitation, i.e., climate change and climate variability \citep{burrell2020,chen2019a,higgins2023shifts,liu2022a,seddon2016,zhu2016}.
While the spatial arrangement of vegetation on a large scale is primarily dictated by climatic factors, the interplay between climate variability and the short-term dynamics of vegetation introduces a higher level of complexity \citep{papagiannopoulou2017, pelletier2015}.
Different Machine Learning (ML) approaches have been suggested to capture the complex nonlinear dynamics of those short-term dynamics. 
However, the employed models are either limited to a specific region \citep{robin2022a,smith2023earthpt} or use a coarse global resolution with one pixel covering at least \SI{0.5}{\degree}($\approx$ \SI{55}{\kilo\metre}) \citep{chen2021,kraft2019identifying}.
While there are statistical approaches that globally quantify the effect of climate variability on vegetation change on a finer spatial resolution up to \SI{0.083}{\degree}, they only consider meteorological data on a coarse time scale, \textit{e.g.}, one data point per month \citep{ burrell2020, seddon2016}.
Nonetheless, the availability of long-term weather reanalysis datasets such as ERA5 \citep{era5}, which comprises hourly high-resolution measurements of \SI{0.25}{\degree} ($\approx$ \SI{27}{\kilo\metre}) per pixel, provide the opportunity to model dependencies of short-term changes in meteorological variables on vegetation activity on a fine spatial and temporal resolution.

Recently, Deep Learning (DL) models have demonstrated the capability to efficiently parse and exploit those vast amounts of meteorological data in the context of medium-range weather forecasting.
Architectural improvements and increased compute availability have led to DL-based weather models that now perform on par with commonly used numerical weather systems \citep{bi2023accurate,lam2023learning,pathak2022fourcastnet}.
These approaches learn a spatial representation of the atmosphere's state by forecasting future atmospheric states.
Previous studies have already shown that their trained atmospheric models can be finetuned to effectively solve other climate-related tasks such as statistical downscaling and climate projections \citep{lessig2023atmorep,nguyen2023climax}.

Based on these advances, this work investigates how the pre-trained weather forecasting model FourCastNet (FCN) \citep{pathak2022fourcastnet} can be adapted for globally modeling the normalized difference vegetation index (NDVI) \citep{tucker1986satellite,vermote2019noaa}, a commonly used index for approximating vegetation activity \citep{ferchichi2022}.
We outline an approach building upon a state-of-the-art DL architecture for processing spatio-temporal data, which enables global modeling of the NDVI at a high spatial (\SI{0.25}{\degree}) and temporal (daily) resolution with a single model.
We investigate how to utilize FCN's atmospheric knowledge by comparing a finetuned model versus a model trained from scratch.
Additionally, we analyze the training time and data needed to make FCN an effective vegetation model in three ablation studies.

\section{Pre-trained weather models for vegetation modeling}

\paragraph{Dataset}
For our study, we use daily global weather data from ERA5~\citep{era5} at a resolution of \SI{0.25}{\degree} (\num{720} x \num{1440} pixel) from the years 1982 to 2013.
We use the same \num{20} predictor variables as \cite{pathak2022fourcastnet}:
zonal and meridional wind velocity (\SI{10}{\metre} above ground, at \SI{1000}{\hecto\pascal}, \SI{850}{\hecto\pascal} and \SI{500}{\hecto\pascal}),
temperature (\SI{2}{\metre} above ground, at \SI{850}{\hecto\pascal} and \SI{500}{\hecto\pascal}),
geopotential (at \SI{1000}{\hecto\pascal}, \SI{850}{\hecto\pascal}, \SI{500}{\hecto\pascal}, and \SI{50}{\hecto\pascal}),
relative humidity (at \SI{850}{\hecto\pascal} and \SI{500}{\hecto\pascal})
surface pressure, mean sea level pressure, and total column water vapor.
One sample has the dimensionality (\num{20} x \num{720} x \num{1440}).
The NDVI data \citep{vermote2019noaa} is our target variable, regridded linearly from originally \SI{0.083}{\degree} to ERA5's \SI{0.25}{\degree} resolution.
This vegetation index is computed from satellite observations as the normalized difference between the spectral reflectances in the near-infrared and red wavebands \citep{tucker1986satellite}.
It ranges from $-1$ to $1$. 
Negative values indicate water, positive values around zero indicate barren land, and values close to one indicate dense vegetation.
NDVI data after 2013 was not considered for this study, as our analysis (\textit{cf.} \cref{fig:ndvi_distribution}) shows a noticeable data shift beginning in 2014

\paragraph{Method}\label{par:method}
To investigate the applicability of pre-trained weather models for globally modelling vegetation activity, we use the FCN Deep Learning model, whose pre-trained weights are publicly available \citep{fcn_weights}.
FCN is a comparatively lightweight weather model (\textit{cf.} \cite{bi2023accurate,chen2023fengwu,lessig2023atmorep,nguyen2023climax})  with a total of \SI{73}{\million} parameters distributed over \num{8} Transformer-like \citep{vaswani2017attention} encoder blocks, see \cref{fig:architecture-overview} for an architectural overview.
Each of these blocks has \SI{5}{\million} parameters and uses an Adaptive Fourier Neural Operator layer \citep{guibas2021adaptive} replacing the attention mechanism 
\citep{bahdanauCB14}.

We adopt the FCN to the NDVI modeling task by replacing the original weather prediction head with a randomly initialized fully-connected layer with the $\tanh$ activation function.
For modeling the effects of short-term climate variability on vegetation activity, FCN is trained on modeling the NDVI for the same timestep as the daily input weather variables.
For finetuning, we initialize the original FCN model with the pre-trained weights, while for training FCN from scratch, we freshly initialize all model weights.

\paragraph{Comparison models}\label{par:baselines}
As a simple baseline, we designed and hyperparameter-optimized a convolutional neural network (CNN) with the details given in \cref{app:baseline_details}.
We further compare us with two recent data-driven models from ecology literature:

The first approach is a global long short-term memory (LSTM) \citep{hochreiter1997long} model by \cite{kraft2019identifying}, trained on single-location time-series of meteorological variables at a \SI{15}{\day} temporal resolution and a \SI{0.5}{\degree} spatial resolution, half of our resolution, with globally shared weights.
This approach reflects the so-called memory effect of vegetation, i.e., that preceding vegetation states can have longer-lasting effects on vegetation activity \citep{dekeersmaecker2015}.
The second data-driven approach trains separate local, weekly state space model (SSM) on \num{100} locations across the globe, guided by equations describing the interplay between the used climate-forcing data \citep{higgins2023shifts}.

\paragraph{Experimental setup}\label{par:exp_setup}
The dataset is split into a training (1982-2010), validation (2011-2012), and test (2013) set.
Training details for FCN can be found in \cref{app:fcn_training_details}.
We perform three individual ablation studies:.
In study I, we vary the number of finetuning epochs between \num{1} to \num{200} epochs.
The number of frozen parameters during finetuning is varied in study II, where we freeze between one and eight of FCN's Transformer blocks in ascending order.
Lastly, in study III, the number of finetuning data is modified by selecting a random \SI{10}{\percent} to \SI{90}{\percent} subset of the training years.

\paragraph{Evaluation}\label{par:eval}
We evaluate all models by computing the root mean squared error (RMSE) and {$\text{R}^2$} score on the test set.
The {$\text{R}^2$} score measures the goodness of fit of the model proportional to the temporal variation of the target NDVI values at a given pixel and ranges from \num{1} (best) to $-\infty$.
For global evaluation in comparison with the LSTM results reported by \cite{kraft2019identifying}, we match their evaluation scheme, thereby removing the same noisy pixels and accounting for varying pixel areas through latitude-weighting (\textit{cf.} \cref{appendix:evaluation}).
Also, our models' outputs and target values are aggregated to \num{15}-day averages to match their temporal resolution.
For local evaluation in comparison with the SSMs results provided by \cite{higgins2023shifts}, we evaluate our model locally at the same \num{100} locations.
At these locations, FCN model outputs and target values are aggregated to \num{7}-day averages as in \cite{higgins2023shifts}, while evaluating the SSMs exclusively on the 2013 test year.

\section{Results and discussion}
\begin{table}[]
    \centering
    \caption{Test year results.
    Left: Latitude weighted global evaluation of the finetuned FCN, FCN trained from scratch, our baseline CNN and the LSTM \citep{kraft2019identifying}.
    Right: Unweighted averages for local evaluation on \num{100} locations for comparison with \cite{higgins2023shifts}.
    $\dag$: local model with global weight-sharing with different variables at \SI{0.5}{\degree} resolution.
    $\ddag$: local models with different variables.
    See \cref{par:eval} and \cref{app:baseline_details} for details.}
    \label{tab:results}
    \sisetup{separate-uncertainty,detect-weight=true,detect-family=trume,detect-inline-weight=math,mode=text}
    
    \begin{tabular}{@{}
        l
        S[table-format=1.4]
        S[table-format=1.4]
        S[table-format=1.4]
        S[table-format=1.3]
        S[table-format=1.4]
        S[table-format=1.4]
        }

    {Model}  & {FCN finetune} & {FCN scratch} & {CNN}  & {LSTM$ \dag  $} & {FCN finetune} & {SSM$ \ddag  $}\\
    \cmidrule(lr){2-5}

    \cmidrule(lr){6-7}
    {Evaluation} & \multicolumn{4}{c} {global, 15-daily} & \multicolumn{2}{c} {local, 7-daily}\\

    \cmidrule(lr){2-5}
    \cmidrule(lr){6-7}
    RMSE    & 0.0403 & 0.0512  & 0.0431 & 0.017 & 0.0547 & 0.0548 \\
    {$\text{R}^2$} &  0.6331 & 0.4977  & 0.6061  & 0.904 & 0.5151 & 0.4038\\

    \end{tabular}
    
\end{table}

\begin{figure}[b]
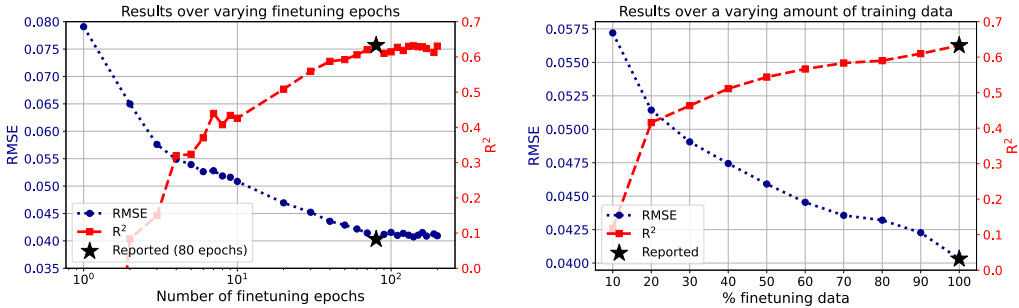

\centering
\begin{subfigure}{.49\textwidth}
\centering
    \includesvg[width=\textwidth]{rmse_r2_epochs.svg}
    \captionlistentry{}\label{fig:var_epochs}
\end{subfigure}
\begin{subfigure}{.49\textwidth}
     \centering
    \includesvg[width=\textwidth]{rmse_r2_varying_training_data.svg}
    \captionlistentry{}\label{fig:var_data}
\end{subfigure}
\caption{Results for ablation studies I \& II. Left: varying number of finetuning epochs. Right: Varying amount of training data.
Results reported in \cref{tab:results} are highlighted in both plots.
}
\end{figure}

\begin{wrapfigure}{R}{0.5\textwidth}
     \centering
    \includesvg[width=0.5\textwidth]{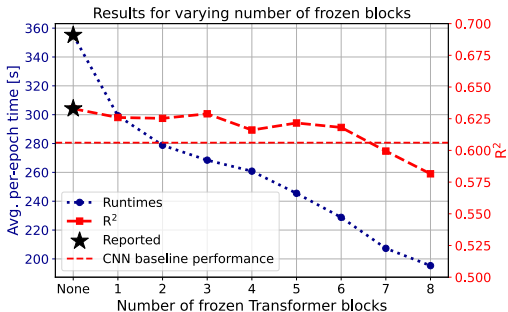}
\caption{Results for ablation study III: varying number of frozen Transformer blocks during finetuning. Runtimes are averaged over five epochs.
}
\label{fig:frozen_blocks}
\end{wrapfigure}
Finetuning the learned atmospheric representation of FCN for vegetation modeling outperforms an NDVI model trained from scratch, as \cref{tab:results} shows.
Here, the scratch model reaches an {$\text{R}^2$} of \num{0.4977} (RMSE: \num{0.0512}).
Finetuning the same model strongly improves NDVI modeling performance up to an {$\text{R}^2$} of \num{0.6331}, which is higher than the strong hyperparameter-optimized CNN baseline with an {$\text{R}^2$} of \num{0.6061} (RMSE: \num{0.0431}).
For the FCN, a parameter search was not considered in this study but could enhance the global ecosystem model further.

To contextualise our performance, the pixel-wise LSTM model by \cite{kraft2019identifying} reaches both the highest {$\text{R}^2$} of \num{0.904} and the lowest RMSE of \num{0.017} but was trained at both lower temporal and spatial resolution than our approach.
Also, as described in \cref{par:baselines}, the time-series approach uses past meteorological data which allows it to model the so-called memory effect of vegetation.
Considering this memory effect seems to be important for modeling the NDVI from weather data and incorporating the respective information into our model might close the observed performance gap compared to the LSTM model.

The average performance on the \num{100} locations selected by \cite{higgins2023shifts} shows that our finetuned FCN reaches a \SI{28}{\percent} higher {$\text{R}^2$} score than the SSMs.
A closer analysis of these locations (see Appendix \cref{tab:extended_location_results}) shows that a single global model can generally learn biome-specific vegetation patterns, but performance is higher in forested regions than in regions with mainly barren land (\textit{cf.} Appendix \cref{fig:r2_global}).
Here, low- to mid-latitude ranges in the Northern Hemisphere are generally well-modelled, while the performance in the Southern Hemisphere and high-latitude regions is worse.
This diminished performance may stem from limited data availability towards the poles.

In ablation study I, performance most strongly rises until \num{80} finetuning epochs,
as shown in \cref{fig:var_epochs}.
Afterwards, performance stagnates at around an {$\text{R}^2$} of \num{0.62}, as more finetuning epochs do not lead to further improvements.

The results for study II in \cref{fig:var_data} show that finetuning FCN on more data improves modeling performance.
The largest jump occurs when doubling the amount of finetuning data from \SI{10}{\percent} to \SI{20}{\percent}.
Beyond, further increases still lead to performance improvements albeit at a slower rate.
Extrapolating \cref{fig:var_data}, this trend suggests that additional data may still enhance performance.

Freezing up to three Transformer blocks in the FCN model results in only minor performance loss, as the results for study III in \cref{fig:frozen_blocks} show.
When more blocks are frozen, the {$\text{R}^2$} scores can drop below the CNN baseline.
However, more frozen blocks reduce the average per-epoch runtime.
It drops from \SI{355}{\second} for the full model to \SI{195}{\second} when finetuning only the newly added vegetation-modeling head.
These results suggest that selectively freezing a moderate number of Transformer blocks can provide a speedup compared to training the full model, while the performance decreases marginally.

\section{Conclusion}
In this work, we investigated how a pre-trained weather model can be adapted for globally modeling vegetation activity as measured by the normalized difference vegetation index.
We finetuned FourCastNet to model the NDVI from \num{20} meteorological variables from the ERA5 dataset and reach a globally averaged test set {$\text{R}^2$} of \num{0.6331}.
This indicates that a weather model finetuned for modeling vegetation activity from high temporal and spatial resolution meteorological data can capture substantial amounts of the NDVI's variability.
Our results further show that training from scratch performs worse than finetuning a pre-trained weather model.
This suggests that during its pre-training phase, the weather model acquires structural knowledge about the atmosphere, which is beneficial for vegetation modeling and which probably is not attained when training from scratch.

While meteorological data partially reflects the impact of climate variability on vegetation, other factors like atmospheric carbon dioxide, soil-related properties and the so-called memory effect are known to be part of the complex interplay between environmental driving forces and vegetation activity \cite{dekeersmaecker2015,piao2020}.
Hence, incorporating further relevant variables into the model while preserving the information in the pre-trained weather models about atmospheric dynamics is an area of future work.
Lastly, we want to highlight that explainable artificial intelligence techniques allow examining attributions of the model's input to its output, such that Deep Learning models can contribute to enhancing our understanding of how globally changing environmental factors affect local ecosystems.

\section*{Acknowledgement}
We are thankful to the HPC team at FAU for providing the computational resources necessary for our research.
This research was conducted in the BigData@Geo 2.0 project which is co-financed by the European Regional Development Fund (ERDF).
The authors gratefully acknowledge the scientific support and HPC resources provided by the Erlangen National High Performance Computing Center (NHR@FAU) of the Friedrich-Alexander-Universität Erlangen-Nürnberg (FAU) under the NHR project ID b214cb. 
NHR funding is provided by federal and Bavarian state authorities.
NHR@FAU hardware is partially funded by the German Research Foundation (DFG) – 440719683.

\bibliography{iclr2024_conference}
\bibliographystyle{iclr2024_conference}

\newpage
\appendix

\section{Extended results and supplementary figures}

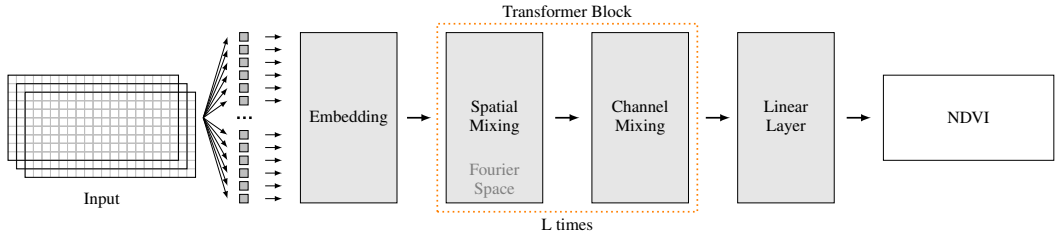
\begin{figure}[h!]
    \centering
    \resizebox{\linewidth}{!}{%
        \begin{tikzpicture}[
      node distance=1cm,
      >=latex,
    ]
    
      \node[draw=black, fill=gray!20, rectangle, minimum height=10em, minimum width=3em, text width=5em, align=center] (embedding) at (0,0) {Embedding};
    
      \foreach \i in {1,...,6} {
        \draw[fill=gray!50] (-2.25cm, {(\i-1)*0.75em - 5em}) rectangle +(0.5em, 0.5em);
        \draw[fill=gray!50] (-2.25cm, {(\i-1)*0.75em + 0.75em}) rectangle +(0.5em, 0.5em);
    
        \draw[-latex, line width = 0.5pt] (-2.25cm + 1.5em, {(\i-1)*0.75em - 4.75em}) -- ++(0.35, 0);
        \draw[-latex, line width = 0.5pt] (-2.25cm + 1.5em, {(\i-1)*0.75em + 1em}) -- ++(0.35, 0);
    
        \draw[-latex, line width = 0.5pt] (-3cm , 0 ) -- (-2.5cm ,{(\i-1)*0.75em - 4.75em});
        \draw[-latex, line width = 0.5pt] (-3cm , 0 ) -- (-2.5cm ,{(\i-1)*0.75em + 1em});
       }
       \draw[dotted, line width=1.5pt] (-2.3cm,-0em) -- (-2cm,-0em);
    
      \node[draw=black, rectangle, minimum height=5em, minimum width=10em, align=center, left=2.5cm of embedding] (input1) {};
    
      \node[draw=black, rectangle, minimum height=5em, minimum width=10em, align=center, left=2.5cm of embedding, shift={(0.5em,-0.5em)}] (input2) {};
    
      \node[draw=black, rectangle, minimum height=5em, minimum width=10em, align=center, left=2.5cm of embedding, shift={(1em,-1em)}] (input3) {};
    
      \node[below=1em] at (input2.south) {Input};

      \begin{scope}[on background layer]
        \draw[step=0.5em, gray!50, very thin, align=center] (input1.south west) grid (input1.north east);
        \draw[step=0.5em, gray!50, very thin, align=center] (input2.south west) grid (input2.north east);
        \draw[step=0.5em, gray!50, very thin, align=center] (input3.south west) grid (input3.north east);    
      \end{scope}

      \node[draw=black, fill=gray!20, rectangle, minimum height=10em, minimum width=3em, right=of embedding, text width=5em, align=center] (spatmix)  {Spatial \\ Mixing};
      \node[above, text width=5em, align=center, text=black!50] at (spatmix.south) {Fourier Space};
    
      \node[draw=black, fill=gray!20, rectangle, minimum height=10em, minimum width=3em, text width=5em, align=center, right=of spatmix] (chanmix) {Channel \\ Mixing};
    
      \node[draw=black, fill=gray!20, rectangle, minimum height=10em, minimum width=3em, text width=5em, align=center, right=of chanmix] (linear) {Linear \\ Layer};
    
      \node[draw=black, rectangle, minimum height=5em, minimum width=10em, align=center, right=of linear] (ndvi) {NDVI};

      \node[draw=orange, dotted, line width=1pt, fit=(spatmix) (chanmix), inner sep=0.5em] (dottedrect) {};
      \node[below] at (dottedrect.south) {L times};
      \node[above] at (dottedrect.north) {Transformer Block};

      \draw[-latex, line width = 1pt] (embedding.east) ++(0.20,0) -- ++(0.5, 0);
      \draw[-latex, line width = 1pt] (spatmix.east) ++(0.25,0) -- ++(0.5, 0);
      \draw[-latex, line width = 1pt] (chanmix.east) ++(0.35,0) -- ++(0.5, 0);
      \draw[-latex, line width = 1pt] (linear.east) ++(0.25,0) -- ++(0.5, 0);
    
    \end{tikzpicture}
    }    
    \caption{Overview of the used architecture based on the FourCastNet model \cite{pathak2022fourcastnet}.
    We initialize the model from pre-trained weights \cite{fcn_weights} and replace the weather-specific head with a linear head for modelling the normalized difference vegetation index (NDVI).}
    \label{fig:architecture-overview}
\end{figure}

\begin{figure}[h!]
    \centering
    \includesvg[width=\textwidth]{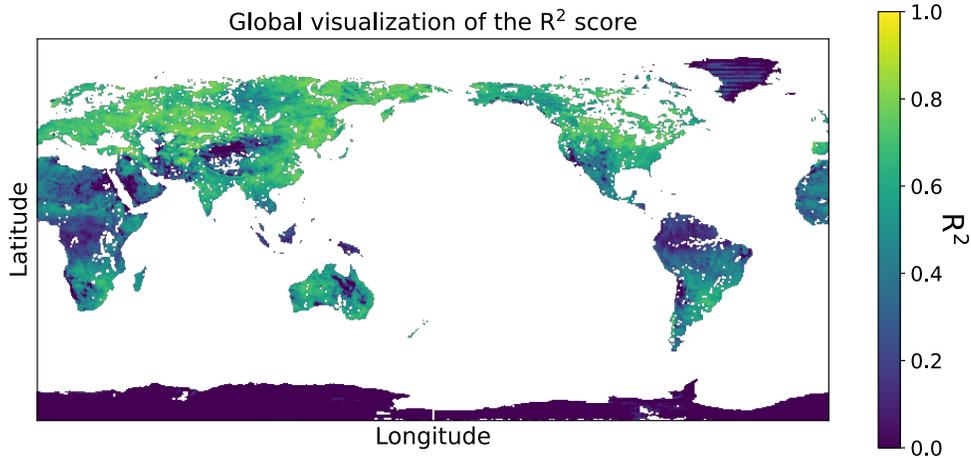}
    \caption{Global visualization of the {$\text{R}^2$} score on the entire test set. {$\text{R}^2$} scores below \num{0} are clipped to 0 for ease of visualization.
    Performance is the strongest for continental Europe and North America and decreases towards higher latitude regions.
    }
    \label{fig:r2_global}
\end{figure}

\begin{table}[h]
    \centering
    \caption{Extended results for local evaluation. Unweighted average of RMSE and $R^2$ score per biome of the \num{100} location from \cite{higgins2023shifts}.
    The finetuned model improves the averaged $R^2$ score across all biomes compared to the SSMs except for boreal forests and tundra. Our global modeling approach thus captures biome-specific NDVI dynamics.
    }
    \sisetup{separate-uncertainty,detect-weight=true,detect-family=trume,detect-inline-weight=math,mode=text,
    round-mode=places,
             round-precision=4,
             table-column-width=3em 
    }
    \begin{tabular}{@{}
        l
        S[table-format=1.4]
        S[table-format=1.4]
        S[table-format=1.4]
        S[table-format=1.4]
        S[table-format=1.3]
        S[table-format=1.4]
        }
    \toprule
    {}  & \multicolumn{2}{c}{RMSE} & \multicolumn{2}{c}{R2} & {Samples}\\
    {Biome} & {FCN} & {SSM} & {FCN} & {SSM} & {} \\
    \midrule
    {Boreal forest} & 0.065645    &  0.083442 &  0.724402  &  0.832095 & {16}\\
    {Grassland} & 0.041639   &   0.044908  & 0.471556  &  0.448353 & {14}\\
    {Mediterranean-type} &  0.036379  &    0.039334 & 0.192081  & -0.600848 & {5}\\
    {Tropical forest} & 0.078943   &   0.050779  & 0.140531 &  -0.051524 & {16}\\
    {Savanna} & 0.051218  &    0.051586 &  0.715146  &  0.662785 & {18}\\
    {Shrubland} & 0.047838  &    0.044490  & 0.302104  &  0.196631 & {16}\\
    {Temperate forest} & 0.059774  &    0.069294  & 0.772599  &  0.524717 & {12} \\
    {Tundra} & 0.042269 &    0.060027 & 0.831170  &  0.914885 & {9}\\
    \bottomrule
    \end{tabular}
    \label{tab:extended_location_results}
\end{table}

\begin{figure}[h]
    \centering
    \includesvg[width=\textwidth]{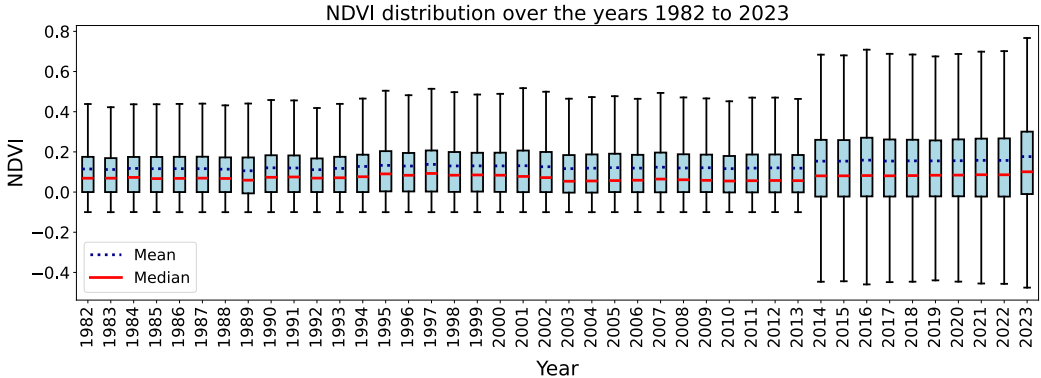}
    \caption{Distribution of the normalized difference vegetation index (NDVI) data from 1982 to 2023.
    NDVI data was only used until 2013 due to the noticeable data shift afterwards.
    }
    \label{fig:ndvi_distribution}
\end{figure}

\newpage

\section{FourCastNet training details}\label{app:fcn_training_details}
We train/finetune the FCN with the Adam optimizer \citep{kingma2014adam} and a learning rate of \num{0.0001} with cosine annealing \citep{loshchilov2016sgdr} for \num{80} epochs (except when varying the number of training epochs in ablation study I, see \cref{par:exp_setup}) using an $l_2$ loss.
Using a binary mask, the loss is only computed for locations with valid NDVI observations.
The weights of the model with the lowest validation loss are kept. We trained our models using a single node equipped with 8 NVIDIA A100 40GB GPUs.

\section{Baseline and comparison models details}\label{app:baseline_details}

\begin{table}[h] 
    \centering
    \caption{Hyperparameter search ranges for the baseline CNN network.}
    \sisetup{
    }
    \begin{tabular}{@{}
    c
    c
    S[table-format=1.0]}
    \toprule
{Hyperparameter} & {Search space} & {Step size}\\
    \midrule
    n\_layers    & 3 to 8 & 1\\
    learning rate & $1\mathrm{e}{-5}$ to $1\mathrm{e}{-5}$ & {log uniform}\\
    epochs & 50 to 100 & 10 \\
    n\_filters & 16 to 512 & 16\\
    kernel\_size & 3 to 7 & 2\\
    \bottomrule
    \end{tabular}
    \label{tab:cnn_param_range}
\end{table}

We use a convolutional neural network (CNN) as a baseline model.
The CNN's hyperparameters were optimized over \num{60} trials on the validation set with the search ranges given in \cref{tab:cnn_param_range}.
The hyperparameter-optimized CNN consists of six convolution layers with \num{64}, \num{16}, \num{512}, \num{128}, \num{512}, \num{128} kernels, respectively.
The used kernel sizes are \num{3}, \num{5}, \num{5}, \num{5}, \num{3}, \num{5}, each with a stride of \num{1}.
The CNN's last layer is a fully connected dense layer with \num{64} neurons, whose output is reshaped to the target resolution of \num{720} x \num{1440}.
The CNN was trained with a learning rate of $0.00006$ for \num{80} epochs.

\section{Evaluation setting}
\label{appendix:evaluation}
\paragraph{Global evaluation}
For global evaluation in comparison to LSTM models by \cite{kraft2019identifying}, we compute 15-days averages of our model output and target values to match their temporal resolution.
To provide a fair comparison to the reported results of \cite{kraft2019identifying}, we replicated their evaluation setting to the best of our knowledge, since their code is not publicly available. 
To remove noisy pixels as defined by \cite{kraft2019identifying}, we remove pixels with \SI{50}{\percent} missing data in the time dimension, pixels with less than \SI{20}{\percent} land mass, and barren-land pixels, which together removes coastal, high-latitude and desert regions.
Further, to account for the varying size of pixels across different latitudes as \cite{kraft2019identifying}, we use latitude-weighted RMSE and {$\text{R}^2$} scores, with the latitude weighting factors $w_1, ..., w_I$ for $I$ evaluated pixels given by
\begin{equation*}
w_i = \frac{\cos(\text{lat}(i))}{\frac{1}{N_\text{lat}}\sum_{j=1}^{N_\text{lat}}\cos(\text{lat}(j))}\quad \forall i \in \{1,...,I\} \quad .
\end{equation*}
We assume that our latitude-weighting is identical to their employed area weighting scheme, \textit{i.e.} given the latitude weights  $w_1, ..., w_I$, the corresponding pixel areas $\mathcal{A}_1, ..., \mathcal{A}_I$ and their total area $A$
\begin{equation*}
\frac{\mathcal{A}_i}{A} = \frac{w_i}{\sum_{j=1}^{I}w_j} \quad \forall i \in \{1,...,I\} \quad .
\end{equation*}
With this assumption, we can show that the reported biome-weighted $\text{RMSE}_\text{global}$ for the biomes $A^1, ..., A^B$ in \cite{kraft2019identifying} is identical to our latitude-weighted $\text{RMSE}$:

\begin{align*}
    \text{RMSE}
    &=\frac{1}{\sum_{j=1}^{I}w_j}\sum_{i=1}^{I}w_i\cdot \text{RMSE}_i\\
    &=\frac{1}{A}\sum_{i=1}^{I}\mathcal{A}_i\cdot \text{RMSE}_i\\
    &=\frac{1}{A}\sum_{b=1}^{B}\sum_{k\in I^b}\mathcal{A}^b_k\cdot \text{RMSE}^b_k \\
    &= \frac{1}{A}\sum_{b=1}^{B}A^b\cdot(\sum_{k\in I^b}\frac{\mathcal{A}^b_k}{A^b}\cdot \text{RMSE}^b_k)\\
    &=\frac{1}{A}\sum_{b=1}^{B}A^b\cdot \text{RMSE}^b\\
    &= \text{RMSE}_\text{global} 
\end{align*}
The above equations also hold for the biome-weighted $R^2_\text{global}$ in \cite{kraft2019identifying} and our latitude-weighted $R^2$ score.
However, we do not perform the same 10-fold spatio-temporal cross-validation due to the long training time of our model. Additionally, note that the LSTM model is evaluated at a coarser resolution of \SI{0.5}\degree than our FCN model (\SI{0.25}\degree) and was trained on a different set of variables. Those are six dynamic meteorological variables and \num{21} static variables including water capacity, water table depth and land cover fractions.

\paragraph{Local evaluation}
For local evaluation in comparison to the state space models (SSM) results provided by \cite{higgins2023shifts}, we average our model output and prediction at the same \num{100} locations to weekly resolution.
We then compute unweighted average RMSE and $R^2$ scores across these locations.
Note that an individual SSM is trained per location, using location-specific air temperature \SI{2}{\metre} above the surface, soil temperature, soil moisture, surface solar radiation, and atmospheric carbon dioxide at \SI{0.083}{\degree} spatial resolution as climate-forcing data.
The target data is the NDVI at a weekly temporal resolution.

\section{Data scaling in ecosystem modeling}\label{app:scaling}
The results for ablation study II, visualized in \cref{fig:var_data}, show that over the evaluated range of number of finetuning data, performance increases as more data is used.
A similar behaviour was observed for earth observation data by \cite{smith2023earthpt}, who trained a pixel-wise autoregressive Transformer model~\citep{radford2019language} on satellite-derived earth observation data.
They noted that scaling their training data also scales model performance.
In lieu of a scaling ``law'' that applies to earth observation data, they assume their model performance to follow the scaling law observed for large language models:
\begin{equation*}
N \sim 20 \; D,
\end{equation*}

where $N$ is the number of model parameters and $D$ is the number of training tokens \citep{hoffmann2022empirical}.
In the following, we assume that this tendency also applies to our vegetation-modeling approach. 

Our training dataset contains \num{29} training years (\num{1982} to \num{2010}) with \num{365} days per years, leading to $29*365=10585$ training samples.
One sample has dimensionality 720 x 1440, and the patch size used to tokenize this image-like input is \num{8}.
For one training sample, this leads to $720/8=90$ tokens covering the latitudinal direction, and $1440/8=180$ tokens for the longitudinal direction, or $90*180=16200$ total tokens for one sample.
Over the entire training data, we thus have $16200*10585=171477000$ tokens in total.

The FourCastNet we use has \SI{73}{\million} parameters.
Assuming that the mentioned scaling law applies, we would thus need to train FCN on $20\;*$ \SI{73}{\million}$\;=\;$\SI{1.46}{\billion} tokens for optimal performance\footnote{Or, alternatively, we should scale down the model to $171477000$ total tokens $/\;20=8.573.850$ parameters, which would roughly be a two-Transformer block model.
The observed results in \cref{fig:frozen_blocks} currently indicate that this configuration does not show the best performance, at least for a setting where these blocks are the only ones being trained, and further blocks -- and thus parameters -- exist but are frozen.
}
However, this calculation assumes the validity of the large language model scaling law.
For autoregressive, image-like data-generating models, different data-scaling might be better suited, such as the ones proposed by \cite{henighan2020scaling}.
Applying these laws to ecosystem models is thus an open research direction.

\end{document}

%% file: math_commands.tex
\usepackage{amsmath,amsfonts,bm}

\def\eqref#1{equation~\ref{#1}}

\def\1{\bm{1}}

\DeclareMathAlphabet{\mathsfit}{\encodingdefault}{\sfdefault}{m}{sl}
\SetMathAlphabet{\mathsfit}{bold}{\encodingdefault}{\sfdefault}{bx}{n}

